\documentclass[sigconf,nonacm]{acmart}

\usepackage{multirow}
\usepackage[utf8]{inputenc}

\usepackage{float}
\usepackage{blindtext}

\usepackage[normalem]{ulem}
\useunder{\uline}{\ul}{}

\usepackage{algorithm,algorithmicx,algpseudocode}

\def\BibTeX{{\rm B\kern-.05em{\sc i\kern-.025em b}\kern-.08emT\kern-.1667em\lower.7ex\hbox{E}\kern-.125emX}}

\begin{document}

%
\title{On-the-Fly Ensemble Pruning in Evolving Data Streams}

%
\author{Sanem Elbasi}
\affiliation{%
  \institution{Bilkent Information Retrieval Group}{Bilkent University}
}
\email{sanem.elbasi@bilkent.edu.tr}

\author{Alican B\"uy\"uk\c{c}ak{\i}r} 
\affiliation{%
  \institution{Bilkent Information Retrieval Group}{Bilkent University}
}
\email{alicanbuyukcakir@bilkent.edu.tr}

\author{Hamed Bonab}
\affiliation{%
  \institution{College of Information and Computer Sciences}{University of Massachusetts Amherst}
}
\email{bonab@cs.umass.edu}

\author{Fazli Can}
\affiliation{%
  \institution{Bilkent Information Retrieval Group}{Bilkent University}
}
\email{canf@cs.bilkent.edu.tr}

%
\renewcommand{\shortauthors}{Elbasi et al.}

%
\begin{abstract}
Ensemble pruning is the process of selecting a subset of component classifiers from an ensemble which performs at least as well as the original ensemble while reducing storage and computational costs. Ensemble pruning in data streams is a largely unexplored area of research. It requires analysis of ensemble components as they are running on the stream, and differentiation of useful classifiers from redundant ones. We present CCRP, an on-the-fly ensemble pruning method for multi-class data stream classification empowered by an imbalance-aware fusion of class-wise component rankings. CCRP aims that the resulting pruned ensemble contains the best performing classifier for each target class and hence, reduces the effects of class imbalance. The conducted experiments on real-world and synthetic data streams demonstrate that different types of ensembles that integrate CCRP as their pruning scheme consistently yield on par or superior performance with 20\% to 90\% less average memory consumption. Lastly, we validate the proposed pruning scheme by comparing our approach against pruning schemes based on ensemble weights and basic rank fusion methods.
\end{abstract}

%
%


%

%

%
\maketitle

\section{Introduction}

Data streams are environments such as network event logs, video streams, call records, and transaction records where vast amounts of data is generated at high speed \cite{aggarwal2007datastreams}. Classification models in data streams have to work under strict time and memory constraints, and be able to adapt changes in the distribution of data over time \cite{bifet2009new}. Such changes in the relation between data points and their corresponding classes over time are called concept drifts \cite{gama2010knowledge}. Robustness against concept drifts is a topic that is extensively studied in the literature \cite{bonab2018goowe, brzezinski2014reacting}. As data continuously arrive, learned information from the past instances become irrelevant under concept drifts. Classification models need to adapt to the new concepts by not only learning the new concepts but also forgetting the now-obsolete ones \cite{brzezinski2014reacting}.

Ensembles are common choices for models in stream environments, as they often improve predictive performance while providing robustness against concept-drifting and time-evolving data \cite{bifet2009new}. Ensembles employ different approaches for dealing with concept drifts, such as replacing the weakest / lowest weight classifier with a new one that is trained with the more recent data \cite{bonab2018goowe}, or updating ensemble weights regularly \cite{wang2003awe, bonab2018goowe}. Despite the popularity and prevalence of ensemble learning in data streams, ensemble pruning is still a vastly undiscovered area of research in stream mining community.

\begin{figure*}
    \centering
    \includegraphics[width=\linewidth]{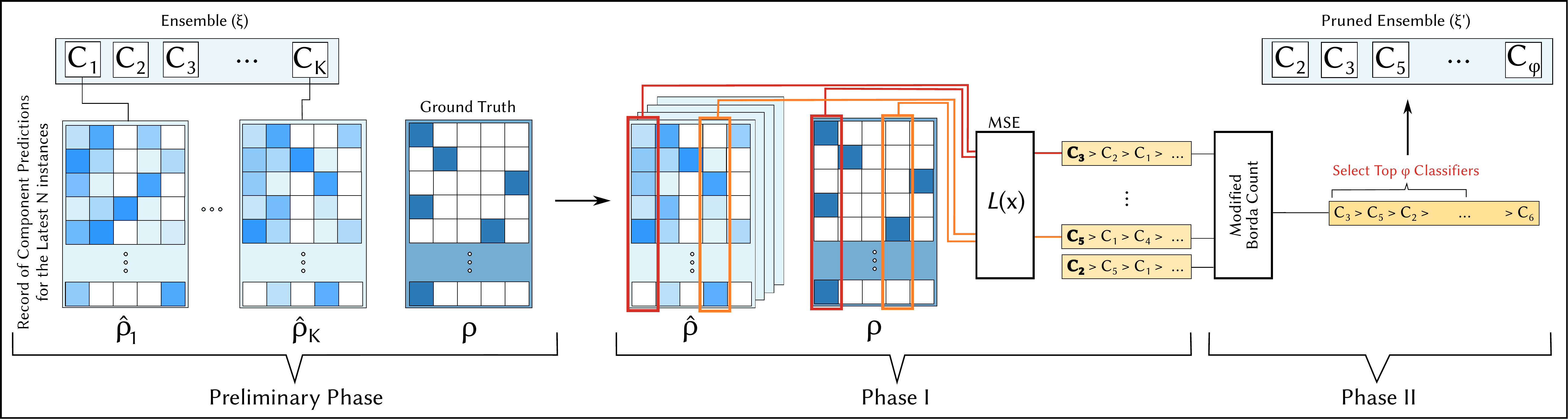}
    \vspace{-0.4cm}
    \caption{Pipeline of Proposed Method. Prediction history of each component in the ensemble are recorded. At each row of records, colored cells indicate relevance scores for corresponding classes. Then, these records are used in calculating class-wise losses for each component, and class-wise rankings of those components. The acquired rankings are combined using a rank fusion algorithm (Modified Borda Count). The resulting top $\varphi$ components ($\varphi < K$) are selected, and the rest is pruned.}
    \label{fig:method-figure}
\end{figure*}

In this study, we tackle the task of ensemble pruning in data streams for multi-class classification. Our contributions are as follows. We propose, 
\begin{itemize}
    \item an explicit pruning technique that can be integrated into any type of streaming ensemble, and called whenever pruning is requested. Our method results in pruned ensembles that are better performing in terms of accuracy and more efficient in terms of memory consumption (as demonstrated in Section \ref{sec:results}).
    \item to the best of our knowledge, the first large-scale \footnote{The previous studies \cite{krawczyk2015one, cruz2019fire} experimented on datasets with number of instances in the scales of hundreds or thousands} on-the-fly ensemble pruning method for streaming data.
    \item a class imbalance-aware pruning method, where a pruned ensemble does not lose its ability to classify rare or less-frequent classes, as a result of our class-wise component prediction analysis and ranking (as described in Section \ref{sec:class-wise-analysis-ranking}).
\end{itemize}

\section{Problem Definition and Notation}

We consider the problem of ensemble pruning within the context of supervised classification in data streams. Data stream $\mathcal{D}$ is a (possibly infinite) sequence of time-ordered data that consists of pairs $(X^{(t)}, y^{(t)})$ where $t$ denotes the arrival time. In multi-class classification, the target $y^{(t)}$ has $L$-many classes where $L>2$. An ensemble model $\xi$ is a group of $K$ component classifiers, i.e. $\xi = \{ C_1, C_2, \dots, C_K \}$ where the prediction of the model for an instance is a combination of individual hypotheses ($h_k (X)$) of its component classifiers where $k$ denotes the classifier index. 

Let $\hat{\rho}_{k,l}$ be the record of the predictions of $k$th component for the $l$th class on a sliding window containing the latest $N$ instances. Similarly, let $\rho_{l}$ be the record of the ground truth information for the $l$th class on the sliding window.

The aim of ensemble pruning is selecting a subset of classifiers $\xi' \subset \xi$ such that the ensemble's both efficiency and predictive performance are improved. Here, let us denote the size of the pruned ensemble as $|\xi'| = \varphi$ where $\varphi < K$. 

\section{Related Work}
\textbf{\textit{Online Ensembles}}. The literature on the online ensembles is abundant \cite{krawczyk2017ensemble}. Here, we only mention the ensembles that are used in our experiments. One of the most commonly used and well-known ensembles in data streams is AWE (Accuracy Weighted Ensemble) \cite{wang2003awe}. It is a chunk-based ensemble where each classifier are weighted according to their expected accuracy in the latest chunk. Similarly, GOOWE \cite{bonab2018goowe} is another chunk-based stacking ensemble that assigns optimal weights to its components using a geometric interpretation of target space and solving an optimization problem as the chunks are processed. Both of these ensembles are dynamic and they replace one of their components at the end of each chunk.


\noindent
\textbf{\textit{Pruning in Data Streams}}. A related topic to on-the-fly ensemble pruning is Dynamic Ensemble Selection (DES) \cite{ko2008dynamic}. In DES, a subset of components from a pool is selected for each test instance. DES assigns estimated levels of competence for each classifier in the pool, so that a selection method selects the classifiers to be assigned to incoming instances. Note that, unlike ensemble pruning, selection of a subset of classifiers for every instance does not reduce the size of the pool of classifiers. Firefly \cite{krawczyk2015one} introduces a swarm intelligence approach to pruning with \textit{one-class} classifiers where input classifiers are members of a firefly population. Interactions between fireflies describe the effectiveness of the classifiers, and best representatives of diverse groups in the population are selected as the pruning step. Lastly, FIRE-DES++ \cite{cruz2019fire} incorporates a pre-selection stage where only the classifiers that can correctly classify at least a pair of instances with different classes can proceed to the next stage of classification. Again, the pre-selection stage in FIRE-DES++ does not change the size of the pool of classifiers.

\vspace{-0.1cm}
\section{Proposed Method}
\begin{algorithm}
    \caption{\texttt{CCRP}: Class-wise Component Ranking-based Pruner}
    \label{alg:ccrp}
    \begin{algorithmic}[1]
        \Require $\mathcal{D}$: data stream, $\xi$: ensemble, $\varphi$: pruned ensemble size
        \Ensure $\xi'$: pruned ensemble
        \State Initialize $\rho$ as an empty FIFO buffer of size $N$.          
        \State Initialize $\hat{\rho}$ as an empty FIFO buffer of size $K \times N$.
        \For{$(X,y) \in \mathcal{D}$}
            \State $\hat{\rho}_k.$ \texttt{append}$(h_k (X)) \; $ for $\forall k$  \Comment{[ Preliminary Phase ]}\label{code:preliminary}
            \State $\rho.$ \texttt{append}$(y)$ 
            
            \If{prune}  \Comment{[ Phase I ]} \label{code:phase-1} 
                \For{$l \in L$}     \Comment{Start CCRP}
                    \For{$k \in K$}
                        \State $ scores.$\texttt{append}$($\texttt{MSE}$(\rho_{l}, \hat{\rho}_{k,l}))$\label{code:mse-line}  \Comment{Class-wise MSEs}
                    \EndFor
                    \State $ ranks[l] =$ \texttt{argsort}$(scores)$ \Comment{Class-wise order of components}
                \EndFor
                \State $ccrp\_ranks =$ \texttt{ModifiedBorda}$(ranks)$ \label{code:ModifiedBorda} \Comment{[ Phase II ]}\label{code:phase-2} 
                \State $\xi' \leftarrow$ top ranked $\varphi$ components based on $ccrp\_ranks$  
                \State $\xi \leftarrow \xi'$ \Comment{Pruned ensemble in effect}
            \EndIf
        \EndFor
    \end{algorithmic}
\end{algorithm}

We introduce, CCRP, \textbf{C}lass-wise \textbf{C}omponent \textbf{R}ankings based \textbf{P}runer for multi-class online ensembles. CCRP can be integrated to work with both dynamic and static multi-class ensembles to improve the performance and memory consumption. CCRP can be performed at any point in time on the stream. The proposed method consists of three phases: \textit{\textbf{Preliminary Phase}}: Recording Component Predictions on the Sliding Window, \textit{\textbf{Phase I}}: On-the-Fly Performance Analysis and Class-wise Ranking of Components, and \textit{\textbf{Phase II}}: Fusion of Rankings and Component Selection (see Figure \ref{fig:method-figure}).


\subsubsection*{Preliminary Phase: Recording Component Predictions on the Sliding Window.}  This phase only applies to the ensembles that are not already recording the component predictions on the sliding window (e.g. OzaBagging \cite{oza2005onlinebb}). In this phase, for the latest $N$ data instances, predictions of classifiers and the ground truth are recorded as  $\rho$ and $\hat{\rho}$ respectively (Alg.\ref{alg:ccrp} line \ref{code:preliminary}). Since only the latest $N$ instances are kept in the records, the required memory for this process is fixed, despite being proportional to the number of classes.

\subsubsection*{Phase I: On-the-Fly Performance Analysis and Class-wise Ranking of Components.} This is the initial phase of CCRP in which per class performances of classifiers are measured. In this phase, for each class $l$, predictions of classifiers and the ground truth are extracted as $\hat{\rho}_{k,l}$ and $\rho_{l}$ respectively (Alg.\ref{alg:ccrp} line \ref{code:phase-1}). Then, scores of classifiers are calculated using Mean Square Error (Eqn. \ref{eq:loss-mse}) for each class (Alg.\ref{alg:ccrp} line \ref{code:mse-line}).
\label{sec:class-wise-analysis-ranking}

\vspace{-0.3cm}
\begin{equation}
\label{eq:loss-mse}    
    \mathcal{L}_{k,l}(X) = \sum^{N}_{i=1} \Big( \hat{\rho}_{k,l}^{(i)} - \rho_{l}^{(i)} \Big)^2, \qquad 1 \leq k \leq K, \; 1 \leq l \leq L
\end{equation}

\subsubsection*{Phase II: Fusion of Rankings and Component Selection.} In the final phase, CCRP generates the overall ranking of the classifiers, using a modified version Borda Count \cite{nuray2006automatic} rank fusion method (Alg.\ref{alg:ccrp} line \ref{code:ModifiedBorda}). In Modified Borda Count (MBC, hereafter), CCRP assigns $K \times L$ points (instead of $K$ points in the regular Borda Count \cite{nuray2006automatic}) to the highest ranking components in the class-wise rankings. This ensures the winning component for each class to appear at the top $L$ places in the overall ranking. Afterwards, the second classifier gets $K - 1$ points and each proceeding classifier gets 1 point less than its successor. Then, the top $\varphi$ classifiers from the overall ranking are selected as the members of the pruned ensemble. It is recommended \cite{bonab2019less} that the pruned ensemble size should be at least equal to the number of classes. Taking this into account, CCRP guarantees that the best performing classifier for each class is included in the pruned ensemble when $\varphi \geq L$.

\section{Experimental Setup and Results} \label{sec:results}

\subsection{Setup}
The proposed method and modifications to existing ensembles are integrated into scikit-multiflow \cite{montiel2018scikit} library. The experiments are evaluated prequentially \cite{gama2009issues}. We report prequential and overall accuracy as performance metrics. In addition, for memory efficiency analysis in the experiments, we define \textit{Mean Memory Consumption Ratio} ($\mu$) which indicates the percentage of average memory the pruned ensemble uses with respect to the original ensemble (Eqn. \ref{eq:mmcr}).

\vspace{-0.3cm}
\begin{equation}
\label{eq:mmcr}
    \mu = \frac{\sum_{\forall \text{chunk}} \; Size(\xi')}{\sum_{\forall \text{chunk}} \; Size(\xi)}
\end{equation}

Throughout the experiments, we use two typical dynamic ensembles (AWE \cite{wang2003awe} and GOOWE \cite{bonab2018goowe}) with Hoeffding Trees \cite{domingos2000mining} as their base classifiers. CCRP is performed whenever the ensembles are full, i.e. when the ensemble size reaches $K$. After pruning, the ensembles continue to grow until they reach the maximum size $K$, where pruning takes place again. This process continues indefinitely.

\begin{table}[!h]
\caption{Datasets}
\vspace{-0.3cm}
\begin{tabular}{l|lrrr}
                                     & \textbf{Name}                                                    & \multicolumn{1}{l}{\textbf{\# Features}} & \multicolumn{1}{l}{\textbf{\# Classes}} & \multicolumn{1}{l}{\textbf{\# Instances}} \\ \hline
\multirow{3}{*}{\rotatebox{90}{\textbf{Real}}}       &  COVTYPE \cite{bifet2013efficient}                                                         & 54                                            & 7                                            & 581.012                                      \\
                                     & Poker Hand \cite{bifet2013efficient}        & 10                                            & 10                                           & 829.201                                      \\
                                     & Rialto    \cite{losing2016knn}                                                       & 27                                            & 10                                           & 82.250                                       \\ \hline
\multirow{3}{*}{\rotatebox{90}{\textbf{Synth}}} & Mov. Squares  \cite{losing2016knn}     & 2                                             & 4                                            & 200.000                                      \\
                                     & Mov. RBF \cite{losing2016knn}        & 10                                            & 5                                            & 200.000                                      \\
                                     & Trn. Chssbrd  \cite{losing2016knn}   & 2                                             & 8                                          & 200.000                                     
\end{tabular}
\label{tab:datasets}
\end{table}

The models are ran on three real-world and three synthetic well-known datasets with concept drifts \cite{bifet2013efficient, losing2016knn}. A summary of the dataset information is provided in Table \ref{tab:datasets}.

\begin{figure}
    \vspace{-0.35cm}
    \centering
    \includegraphics[width=\linewidth]{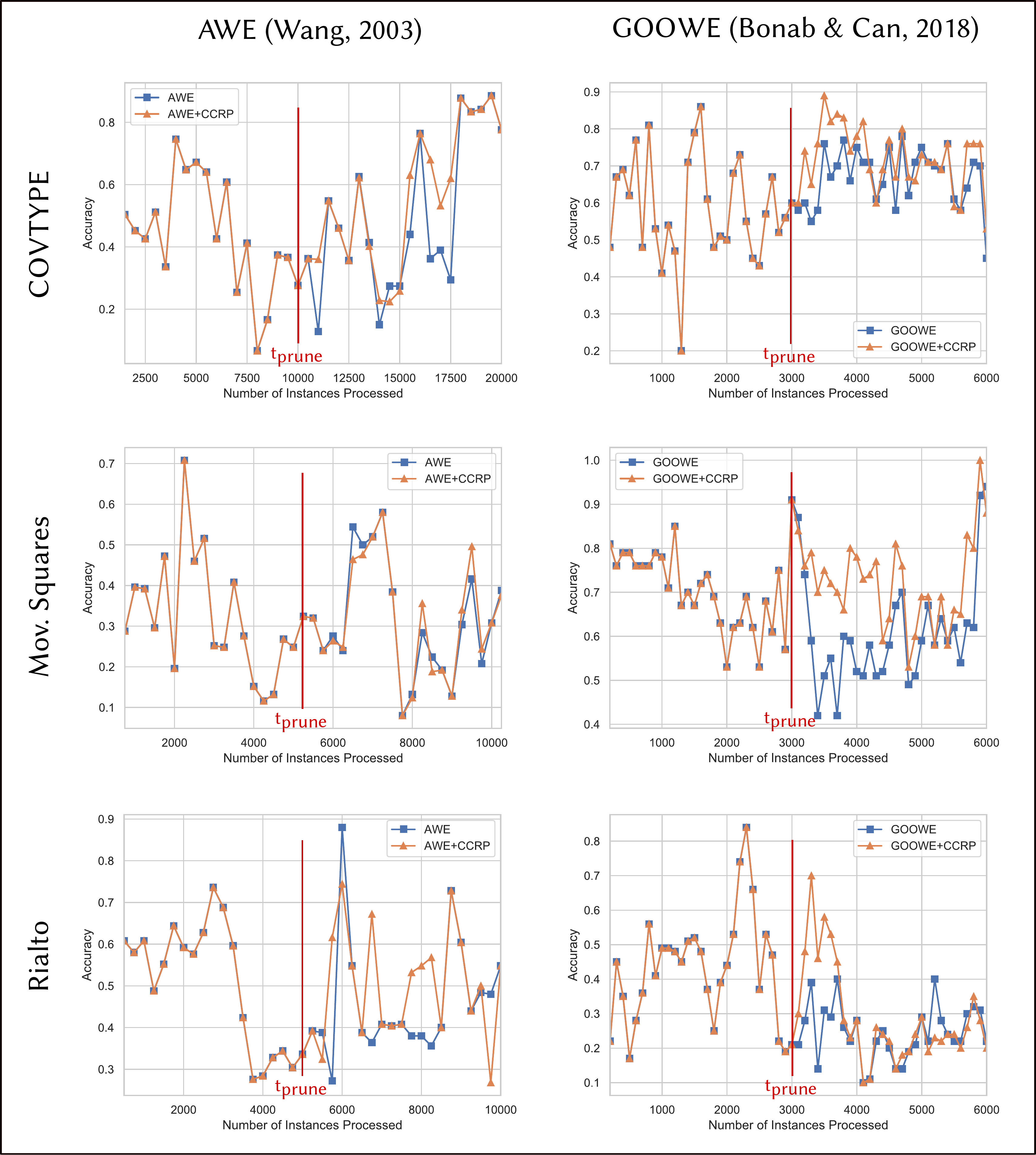}
    \caption{The impact of CCRP ($\varphi = L$) on the prequential accuracy of two ensemble models on three datasets over the first few thousand instances. The first occurrence of pruning is denoted with a vertical red line.}
    \label{fig:ccrp-with-different-ensembles}
\end{figure}

\subsection{Effect of CCRP on Predictive Performance and Memory Efficiency}
\textit{Does CCRP yield more effective and efficient ensemble models?} We respond to this question by taking both accuracy and memory consumption ($\mu$) into account. We perform experiments with AWE (where $K = 20$) and GOOWE ($K = 30$) to investigate the effect of CCRP on the ensembles. 

To examine the change in the behavior of ensembles, \textit{ceteris paribus}, we consider the case when CCRP is performed for the first time on the stream. In Figure \ref{fig:ccrp-with-different-ensembles}, it can be observed that the overall accuracy of the pruned ensemble is always higher than the original ensemble \textit{after} pruning with CCRP. In addition to the improvement in accuracy, it should be noted that the ensembles after pruning are relatively more robust to the changes in the data stream. This can be observed from the steepness of the declining accuracies of the original ensembles compared to the pruned ensembles.

\begin{table}[]
\caption{Effect of CCRP on Accuracy and Mean Memory Consumption Ratio ($\mu$) for Different Ensembles}
\vspace{-0.3cm}
\begin{tabular}{l|cc|c|cc|c}
\multicolumn{1}{c|}{} & \multicolumn{3}{c|}{\textbf{AWE} \cite{wang2003awe}}                           & \multicolumn{3}{c}{\textbf{GOOWE} \cite{bonab2018goowe}}                         \\ \cline{2-7} 
                      & \multicolumn{1}{c}{Default} & \multicolumn{1}{c|}{CCRP} & $\mu$ & \multicolumn{1}{c}{Default} & \multicolumn{1}{c|}{CCRP} & $\mu$ \\ \hline
COVTYPE               &    {\ul 0.661}     &         0.659                  &  \textbf{75\%} &        {\ul 0.852}          &          0.829     &  \textbf{10\%} \\
PokerHand             &    0.525     &        {\ul 0.585}          & \textbf{83\%}  &          {\ul 0.712}       &            0.651       &  \textbf{14\%} \\
Rialto                &       0.377        &      {\ul 0.379}          &  \textbf{80\%}  &       {\ul  0.428}           &      0.420                &  \textbf{55\%} \\
Mov. Squares          &       0.478        &       {\ul 0.532}         &  \textbf{71\% } &        0.354            &       {\ul 0.359}     &  \textbf{52\%} \\
Mov. RBF              &       {\ul 0.499}     &      {\ul 0.499}         &  \textbf{72\%}  &        {\ul 0.433}     &         0.396         &  \textbf{19\%} \\
Trn. Chssbrd       &      {\ul 0.146}   &        0.141               &  \textbf{71\%}  &        0.685 &             {\ul 0.791}         &  \textbf{70\%}
\vspace{-0.4cm}
\end{tabular}
\label{table:acc-memory}
\end{table}

Figure \ref{fig:ccrp-with-different-ensembles} indicates that the initial calls of CCRP \textit{locally} improves the ensemble performance. Yet, \textit{is that the case throughout the stream?} To inspect the effect of CCRP on both accuracy and memory consumption throughout the stream, we experiment with AWE ($K=20$) and GOOWE ($K=30$) on all of the datasets, and report overall accuracies, as well as $\mu$ (Table \ref{table:acc-memory}). 

In Table \ref{table:acc-memory}, notice that CCRP yields $10\% \leq \mu \leq 83\%$, which means the pruned ensembles consume at most $90\%$ and at least $17\%$ less memory over the course of their run. In 5 out of 12 cases, CCRP improves the performance of ensembles while reducing the memory consumption by $20$ to $25\%$. In a small number of cases (3 out of 12: GOOWE on COVTYPE, PokerHand and MovingRBF), ensembles without pruning outperforms the ones with pruning. These are also the cases where CCRP yields the most drastic reductions in memory consumption, around $80$ to $90\%$. This behaviour may occur due to removing old and strong components of the ensemble in the presence of drifts. For the remaining 4 cases, the overall accuracy for the pruned models are on par with the ones that are not pruned. A decrease in memory consumption around $30\%$ can be observed in these cases, as well. These memory savings imply similar savings in CPU time, since memory consumption of an ensemble is proportional to the complexity of its components.

\subsection{Superiority of CCRP over Different Baseline Selection Techniques}
\textit{Is using CCRP's pruning scheme indeed better than selecting the highest weighted $\varphi$ components from a weighted ensemble? What difference does using MBC make with respect to the regular Borda Count?} To investigate these questions, we design the following experiment: We choose a weighted ensemble (GOOWE \cite{bonab2018goowe}) which normally employs a replacement policy where the lowest weight component is replaced at the end of each chunk. We, then, modify GOOWE's component replacement policy so that it removes its component using CCRP, and Borda Count with $\varphi = K - 1$, and places the newly trained component to the freed-up spot. Overall accuracy values for several datasets are reported in Table \ref{table:differentPrune}. The winning method(s) for each dataset is underlined.
 
\begin{table}[!h] 
\caption{Accuracy of GOOWE (with $K = 10$) using Different Pruning Schemes (with $\varphi=9$) for the Selected Datasets}
\vspace{-0.3cm}
\begin{tabular}{l|ccc}
             & \textbf{CCRP} & \textbf{Weight-based} & \textbf{Regular Borda} \\ \hline
COVTYPE      & {\ul 0.839}              & 0.796                      & 0.703                     \\
Poker        & {\ul 0.726}               & 0.667                      & 0.701                     \\
Mov. RBF     & {\ul 0.543}              & 0.537                      & {\ul 0.543}                    \\
Trn. Chessboard     & {\ul 0.452}              & 0.450                      & 0.449                    
\end{tabular}
\vspace{-0.3cm}
\label{table:differentPrune}
\end{table}

This experiment shows that the internal component weights of an ensemble do not necessarily indicate their contribution to the overall performance of the ensemble. In addition, CCRP's superiority over using Regular Borda \cite{nuray2006automatic} shows MBC's capability of handling class imbalance. By assigning $K \times L$ points to the best performing classifier for a less frequent class, MBC ensures that that classifier is not ranked poorly and selected for the resulting subset of classifiers.

\vspace{-0.3cm}
\section{Conclusion and Future Work}
In this work, we present an on-the-fly ensemble pruner for evolving data streams for the first time in the literature. Our approach utilizes predictions of ensembles to generate class-wise rankings and combine these rankings with a rank fusion algorithm. We demonstrate that the proposed pruning scheme yields smaller and more efficient ensembles with on par or better predictive performance consistently. Additionally, we show that the proposed rank fusion algorithm works better than its alternatives. Our experiments on different ensembles provide evidence that our method can be integrated into any streaming ensemble.

For future work, we plan to do a rigorous analysis of diversity of the ensemble components before and after pruning, considering various diversity measures \cite{krawczyk2017ensemble}. The proposed approach can be extended by incorporating diversity of components into the rank fusion process. Additionally, "how to select the optimal prune size ($\varphi$)" is still an open question. The effect of ensemble parameters and dataset properties on $\varphi$ can be examined. Lastly, we observed that the time of the pruning is crucial to obtain high predictive performance. Therefore, integrating a concept drift detector into the proposed method and pruning the ensemble whenever a drift is detected may further improve the performance of models.
\vspace{-0.3cm}
%
\bibliographystyle{ACM-Reference-Format}
\bibliography{sample-base}
\vspace{-0.3cm}

%

\end{document}